\documentclass[11pt,a4paper]{article}
\usepackage{times,latexsym}
\usepackage{url}
\usepackage[T1]{fontenc}
\usepackage{enumitem}
\usepackage{amsmath}
\usepackage{graphicx}
\usepackage{multirow}
\usepackage{cases}
\usepackage{mathtools}
\usepackage{CJKutf8}
\usepackage{booktabs}
\newcommand{\tabincell}[2]{\begin{tabular}{@{}#1@{}}#2\end{tabular}} 
\newcommand{\modi}{\textcolor{black}}
\newcommand{\fimodi}{\textcolor{black}}
\usepackage[acceptedWithA]{tacl2021v1}

\title{A Survey on Cross-Lingual Summarization}

\author{Jiaan Wang\textsuperscript{1}\thanks{ \ \ Work was done when Jiaan Wang was interning at Pattern Recognition Center, WeChat AI, Tencent Inc, China.}, \ Fandong Meng\textsuperscript{2}\thanks{ \ \ Corresponding authors.}, \ Duo Zheng\textsuperscript{4}, \ Yunlong Liang\textsuperscript{2}\\
\bf {Zhixu Li\textsuperscript{3}\footnotemark[2], \ Jianfeng Qu\textsuperscript{1} and \ Jie Zhou\textsuperscript{2}} \\
\small{\textsuperscript{1}School of Computer Science and Technology, Soochow University, Suzhou, China} \\
\small{\textsuperscript{2}Pattern Recognition Center, WeChat AI, Tencent Inc, China} \\
\small{\textsuperscript{3}Shanghai Key Laboratory of Data Science, School of Computer Science, Fudan University, Shanghai, China} \\
\small{\textsuperscript{4}Beijing University of Posts and Telecommunications, Beijing, China} \\
\small \texttt{jawang1@stu.suda.edu.cn}, \texttt{\{fandongmeng,yunlonliang,withtomzhou\}@tencent.com} \\
\small \texttt{zd@bupt.edu.cn},
\texttt{zhixuli@fudan.edu.cn},
\texttt{jfqu@suda.edu.cn}
}

\date{}

\begin{document}
\maketitle
\begin{abstract}
Cross-lingual summarization is the task of generating a summary in one language (e.g., English) for the given document(s) in a different language (e.g., Chinese).
% Cross-lingual summarization aims to summarize the given document(s) in one language (e.g., English) into another language (e.g., Chinese), which is extremely challenging since it requires both the abilities to translate and summarize.
% Cross-lingual summarization converts document(s) in one language (e.g., English) to a summary in another one (e.g., Chinese), which is extremely challenging since it requires both the abilities to translate and summarize.
% 
Under the globalization background, this task has attracted increasing attention of the computational linguistics community.
Nevertheless, there still remains a lack of comprehensive review for this task. Therefore, we present the first systematic critical review on the datasets, approaches\modi{,} and challenges in this field. Specifically, we carefully organize existing datasets and approaches according to different construction methods and solution paradigms, respectively. For each type of datasets or approaches, we thoroughly introduce and summarize previous efforts and further compare them with each other to provide deeper analyses.
% Then, we introduce and summarize each type of datasets or approaches, and further compare them with each other to provide deeper analyses.
%
In the end, we also discuss promising directions and offer our thoughts to facilitate future research.
This survey is for both beginners and experts in cross-lingual summarization, and we hope it will serve as a starting point as well as a source of new ideas for researchers and engineers interested in this area.
\end{abstract}

\section{Introduction}

To help people efficiently grasp the gist of documents in a foreign language, Cross-Lingual Summarization (XLS) aims to generate a summary in the target language from the given document(s) in a different source language.
% As the example shown in Figure x, most information of the given English document is distilled into a shorter Chinese summary by XLS systems.
% The practical benefits of this task are twofold: xxx.
This task could be regarded as a combination of monolingual summarization (MS) and machine translation (MT), both of which are unsolved natural language processing (NLP) tasks and have been continuously studied for decades~\cite{Paice1990ConstructingLA,brown-etal-1993-mathematics}.
\modi{XLS is an extremely challenging task: (1) from the perspective of data, unlike MS, naturally occurring documents in a source language paired with the corresponding summaries in different target languages are rare, making it difficult to collect large-scale and human-annotated datasets~\cite{ladhak-etal-2020-wikilingua,perez-beltrachini-lapata-2021-models}; (2) from the perspective of models, XLS requires both the abilities to translate and summarize, which makes it hard to generate accurate summaries by directly conducting XLS~\cite{cao-etal-2020-jointly}.}
% \modi{Therefore, XLS requires both the abilities to translate and summarize, which makes it a challenging task~\cite{cao-etal-2020-jointly}.}
% \modi{Therefore, it is extremely challenging for the model to directly conduct XLS since it requires both the abilities to translate and summarize~\cite{cao-etal-2020-jointly}.}

% Building XLS systems can help people efficiently grasp the gist of documents in a foreign language and further facilitate information dissemination across the world.
%
\modi{Despite its importance}, XLS has attracted a little attention~\cite{Leuski2003CrosslingualCE,wan-etal-2010-cross} in the statistical learning era due to its difficulties and the scarcity of parallel corpus.
Recent years have witnessed the rapid development of neural networks, especially the emergence of pre-trained encoder-decoder models~\cite{Zhang2020PEGASUSPW,Raffel2020ExploringTL,lewis-etal-2020-bart,Liu2020MultilingualDP,Tang2020MultilingualTW,Xue2021mT5AM}, making neural summarizers and translators achieve impressive performance.
Meanwhile, creating large-scale XLS datasets has proven feasible by utilizing existing MS datasets~\cite{zhu-etal-2019-ncls,Wang2022ClidSumAB} or Internet resources~\cite{ladhak-etal-2020-wikilingua,perez-beltrachini-lapata-2021-models}.
\modi{The aforementioned} successes have laid the foundation for the XLS research field and gradually attracted interest in XLS. In particular, recent researchers put their efforts into solving XLS task and published more than 20 papers over the past five years.
% Recent years have witnessed increased interest in XLS thanks to the emerging of pre-trained multi-lingual encoder-decoder models~\cite{Liu2020MultilingualDP,Tang2020MultilingualTW,Xue2021mT5AM} and the availability of large-scale datasets~\cite{ladhak-etal-2020-wikilingua,perez-beltrachini-lapata-2021-models,Wang2022ClidSumAB}.
%
Nevertheless, there still lacks a systematic review of progresses, challenges and opportunities of XLS.

To fill the above gap and help new researchers, in this paper, we provide the first comprehensive review of existing efforts relevant to XLS and give multiple promising directions for future research.
Specifically, we first briefly introduce the formal definition and evaluation metrics of XLS (\S~\ref{sec:background}), which serves as a strong background before delving further into XLS.
Then, we provide an exhaustive overview of existing XLS research datasets (\S~\ref{sec:datasets}).
%which are divided into synthetic datasets and multi-lingual website datasets according to the different construction methods. 
\modi{In detail, to alleviate the scarcity of XLS data, previous work resorts to different ways to construct large-scale benchmark datasets, which are divided into synthetic datasets and multi-lingual website datasets.}
The synthetic datasets~\cite{zhu-etal-2019-ncls,bai-etal-2021-cross,Wang2022ClidSumAB} are constructed through (manually or automatically) translating the summaries of existing MS datasets from a source language to target languages while the multi-lingual website datasets~\cite{nguyen-daume-iii-2019-global,ladhak-etal-2020-wikilingua,fatima-strube-2021-novel,perez-beltrachini-lapata-2021-models} are collected from websites that provide multi-lingual versions for their content.

\modi{Next}, we thoroughly introduce and summarize existing models, which are organized with respect to different paradigms, i.e., pipeline (\S~\ref{sec:pipeline}) and end-to-end (\S~\ref{sec:e2e}). In detail, the pipeline models adopt either translate-then-summarize approaches~\cite{Leuski2003CrosslingualCE,Boudin2011AGA,wan-2011-using,yao-etal-2015-phrase,Zhang2016AbstractiveCS,Pontes2018CrossLanguageTS,Wan2018CrosslanguageDS,ouyang-etal-2019-robust} or summarize-then-translate approaches~\cite{Orasan2008EvaluationOA,wan-etal-2010-cross}. \modi{In this manner, the pipeline models avoid conducting XLS directly, thus bypassing the model challenge we discussed previously. However, the pipeline method suffers from error propagation and recurring latency, making it not suitable for the real-world scenario~\cite{ladhak-etal-2020-wikilingua}. Consequently, the end-to-end method has attracted more attention. To alleviate the model challenge, it generally utilizes the related tasks (e.g., MS and MT) as auxiliaries or resorts to external resources.}
% The former is more popular since the summarizers can take bilingual information into account, but it is less cost-effective (translating the whole documents instead of summaries) and depends on more powerful MT models to reduce error propagation.
%
%
\modi{The end-to-end models} mainly fall into four categories, i.e., multi-task methods~\cite{zhu-etal-2019-ncls,Takase2020MultiTaskLF,cao-etal-2020-jointly,bai-etal-2021-cross,Liang2022AVH}, knowledge-distillation methods~\cite{Ayana2018ZeroShotCN,duan-etal-2019-zero,Nguyen2021ImprovingNC}, resource-enhanced methods~\cite{zhu-etal-2020-attend,Jiang2022ClueGraphSumLK} and pre-training methods~\cite{dou-etal-2020-deep,xu-etal-2020-mixed,Ma2021DeltaLMEP,chi-etal-2021-mt6,Wang2022ClidSumAB}.
% (1) The multi-task learning methods~\cite{zhu-etal-2019-ncls,Takase2020MultiTaskLF,dou-etal-2020-deep,cao-etal-2020-jointly,bai-etal-2021-cross,Liang2022AVH} utilize the XLS and its related task(s) (e.g., MT or MS) to jointly train a unified model. In this way, the model could also benefit from other tasks and finally improve the performance on XLS;
% (2) The knowledge-distillation-based methods~\cite{Ayana2018ZeroShotCN,duan-etal-2019-zero,Nguyen2021ImprovingNC} regard the XLS model as a student model and further design other models, called teacher models, to learn from the related tasks. And then, the soft labels (e.g., hidden representation or final logits) provided by the teacher models and the primary XLS labels jointly supervise the student model;
% In this way, the implied knowledge of related tasks is distilled into the student model by the teacher model;
% (3) The pre-training-based methods~\cite{Ma2021DeltaLMEP,chi-etal-2021-mt6,Wang2022ClidSumAB,xu-etal-2020-mixed} first use the related tasks and/or self-supervised pretext tasks (e.g., text infilling or span corruption) to pre-train an encoder-decoder model and then fine-tune it on the target XLS task.
% For each category of end-to-end models
For each category, we will thoroughly go through the previous work and discuss the corresponding pros and cons.
Finally, we also point out multiple promising directions on XLS to push forward the future research (\S~\ref{sec:prospects}), followed by the conclusion (\S~\ref{sec:conclusion}).
% Finally, we conclude the paper (\S~\ref{sec:conclusion}).
% To sum up, our contributions are as follows: (1) To the best of our knowledge, this survey is the first that presents a thorough review of XLS; (2) We comprehensively review the existing XLS work and carefully organize them according to different frameworks; (3) We suggest multiple promising directions to facilitate future research.
%
Our contributions are concluded as follows:
\begin{itemize}[leftmargin=*,topsep=0pt]
\setlength{\itemsep}{0pt}
\setlength{\parsep}{0pt}
\setlength{\parskip}{0pt}
    \item To the best of our knowledge, this survey is the first that presents a thorough review of XLS.
    \item We comprehensively review the existing XLS work and carefully organize them according to different frameworks.
    \item We suggest multiple promising directions to facilitate future research on XLS.
\end{itemize}

% The rest of the survey is organized as follows: Section~\ref{sec:background} gives an brief background to the formal definition and evaluation metrics of XLS. In Section~\ref{sec:datasets}, we introduce the existing XLS datasets. Then, we review the pipeline XLS models in Section~\ref{sec:pipeline} and discuss the end-to-end models in Section~\ref{sec:e2e}. We also suggest multiple promising future directions and highlight their challenges in Section~\ref{sec:prospects}. Finally, we conclude the paper in Section~\ref{sec:conclusion}.

\section{Background}
\label{sec:background}

\subsection{Task Definition}
Given a collection of documents in the source language $\mathcal{D}=\{D_{i}\}^{m}_{i=1}$ ($m$ denotes the number of documents and $m\ge$1), the goal of XLS is to generate the corresponding summary in the target language $Y=\{y_{i}\}^{n}_{i=1}$ with $n$ words. The conditional distribution of XLS models is:
\begin{equation}
\nonumber
p_{\theta}(Y|\mathcal{D}) = \prod_{t=1}^{n}p_{\theta}(y_{t}|\mathcal{D},y_{1:t-1})
\end{equation}
where $\theta$ represents model parameters and $y_{1:t-1}$ is the partial \modi{ground truth} summary.

It is worth noting that: (1) when $m$$>$$1$, this task is upgraded to cross-lingual multi-document summarization (XLMS) which has been discussed by some previous studies~\cite{Orasan2008EvaluationOA,Boudin2011AGA,Zhang2016AbstractiveCS}; (2) when the given documents are dialogues, the task becomes cross-lingual dialogue summarization (XLDS) which has been recently proposed by~\citet{Wang2022ClidSumAB}. The XLMS and XLDS are also within the scope of this survey.
Furthermore, we define the source and the target languages in XLS should be two \textit{exactly distinct human languages}, which also means (1) if the source language is in code-mixed style of two natural languages (e.g., Chinese and English), the target language should not be either of the both; (2) the programming languages (e.g., \textsc{Python} or \textsc{Java}) should not be the source or the target language\footnote{If the source language is a programming language while the target language is a human language, the task becomes code summarization which is beyond the scope of this survey.}.
% \footnote{If the source language is a programming language while the target language is a human language, this task becomes code summarization.}

\subsection{Evaluation}
Following MS, ROUGE scores~\cite{Lin2004ROUGEAP} are universally adopted as the basic automatic metrics for XLS, especially the F1 scores of ROUGE-1, ROUGE-2 and ROUGE-L which measure the unigram, bigram and longest common sequence between the ground truth and the generated summaries, respectively. Nevertheless, the original ROUGE scores are specifically designed for English. To make these metrics suitable for other languages, some useful toolkits are \modi{released}, e.g., \texttt{multi-lingual ROUGE}\footnote{\url{https://github.com/csebuetnlp/xl-sum/tree/master/multilingual_rouge_scoring}} and \texttt{MLROUGE}\footnote{\url{https://github.com/dqwang122/MLROUGE}}.
In addition to these metrics based on lexical overlap, recent work proposes new metrics based on the semantic similarity (token/word embeddings), such as \texttt{MoverScore}\footnote{\url{https://github.com/AIPHES/emnlp19-moverscore}}~\cite{zhao-etal-2019-moverscore} and \texttt{BERTScore}\footnote{\url{https://github.com/Tiiiger/bert_score}}~\cite{Zhang2020BERTScoreET} which have been shown their great consistency with the human judgements on MS~\cite{koto-etal-2021-evaluating}.
%
% Human evaluation has also been conducted by some of previous work.

\section{Datasets}
\label{sec:datasets}
% In this section, we introduce existing XLS datasets which are divided into synthetic datasets (\S~\ref{subsec:syn_dataset}) and multi-lingual website datasets (\S~\ref{subsec:mul_datasets}) according to the different construction methods.
% take stock of publicly
In this section, we \modi{review} available large-scale XLS datasets\footnote{There are also some XLS datasets in the statistical learning era, e.g., multiple MultiLing datasets~\cite{giannakopoulos-2013-multi,giannakopoulos-etal-2015-multiling} and translated DUC2001 dataset~\cite{wan-2011-using}. However, these datasets are either not public or extremely limited in scale (typically less than 100 samples). Thus, we do not go into these datasets in depth.} and further divide them into two categories: synthetic datasets (\S~\ref{subsec:syn_dataset}) and multi-lingual website datasets (\S~\ref{subsec:mul_datasets}). 
% Furthermore, we compare the key characteristics of datasets belonging to the same category to provide a deeper understanding.
For each category, we will introduce the construction details and the key characteristics of the corresponding datasets. 
In addition, we compare these two categories to provide a deeper understanding (\S~\ref{subsec:comparision_dataset}).
%  Note that there are some early XLS datasets in the statistical learning era, e.g., multiple MultiLing datasets~\cite{giannakopoulos-2013-multi,giannakopoulos-etal-2015-multiling} and partially translated DUC datasets~\cite{wan-2011-using}. However, these early datasets are either not publicly available or limited in scale.

\subsection{Synthetic Datasets}
\label{subsec:syn_dataset}
Intuitively, one straightforward way to build XLS datasets is directly translating the summaries of a MS dataset from their original language to different target languages.
The datasets built in this way are named synthetic datasets, which could benefit from existing MS datasets.
% Early work xxx.

\vspace{0.5ex}
\noindent \textbf{Datasets Construction.} En2ZhSum~\cite{zhu-etal-2019-ncls} is constructed through utilizing a sophisticated MT service\footnote{\url{http://www.zkfy.com/}} to translate the summaries of CNN/Dailymail~\cite{Hermann2015TeachingMT} and MSMO~\cite{zhu-etal-2018-msmo} from English to Chinese.
% The same as En2ZhSum
In the same way, Zh2EnSum~\cite{zhu-etal-2019-ncls} is built through translating the summaries of LCSTS~\cite{hu-etal-2015-lcsts} from Chinese to English.
Later, \citet{bai-etal-2021-cross} propose En2DeSum through translating the English Gigaword\footnote{LDC2011T07} to German using the WMT’19 English-German winner MT model~\cite{ng-etal-2019-facebook}.
% Though these datasets make great contributions and promote the XLS research to some extent, most of their samples are automatically constructed without human post-editing, resulting in limited quality.
% 
% construct the first XLS benchmark dataset towards dialogues. They

More recently, \citet{Wang2022ClidSumAB} construct XSAMSum and XMediaSum, which directly employ professional translators to translate summaries of two dialogue-oriented MS datasets, i.e., SAMSum~\cite{gliwa-etal-2019-samsum} and MediaSum~\cite{zhu-etal-2021-mediasum}, from English to both German and Chinese.
In this way, their datasets achieve much higher quality than those automatically constructed ones.

\begin{table}[t]
  \centering
  \setlength{\belowcaptionskip}{-10pt}
  \resizebox{0.47\textwidth}{!}
  {
    \begin{tabular}{lcccccc}
    \hline
     \multicolumn{1}{c}{\multirow{2}{*}{\textbf{Dataset}}} &  \multicolumn{1}{c}{\multirow{2}{*}{\textbf{Trans.}}} & \multicolumn{1}{c}{\multirow{2}{*}{\textbf{\modi{Genre}}}} & \multicolumn{1}{c}{\multirow{2}{*}{\textbf{Scale}}} &  \textbf{Src}   & \textbf{Tgt}  \\
     \multicolumn{1}{c}{} & \multicolumn{1}{c}{} & \multicolumn{1}{c}{} & \multicolumn{1}{c}{} & \textbf{Lang.} & \textbf{Lang.}  \\ \hline
    % \multirow{5}{*}{\rotatebox[origin=c]{90}{synthetic} $\begin{dcases} \\ \\ \\ \\ \end{dcases}$}
    En2ZhSum    & Auto. & News  & 371k  & En    &   Zh      \\
    Zh2EnSum    & Auto. & News  & 1.7M  & Zh    &   En      \\
    En2DeSum    & Auto. & News  & 438k  & En    &   De      \\
    % MSAMSum     & Auto. & Dial. & 6k    & En    &   6       \\
    XSAMSum     & Manu. & Dial. & 16k$\times$2   & En    &   De/Zh   \\
    XMediaSum   & Manu. & Dial. & 40k$\times$2   & En    &   De/Zh   \\ \hline
    \end{tabular}
  }
  %  which are organized into synthetic datasets and multi-lingual website datasets
  \caption{Overview of existing synthetic XLS datasets. ``\textit{Trans.}'' indicates the translation method (automatic or manual) to construct datasets. The ``\textit{\modi{genres}}'' of these datasets are divided into news articles and dialogues according to the basic MS datasets. For ``\textit{scale}'', some datasets contain two cross-lingual directions, thus we use $\times$2 to calculate the overall scale. ``\textit{Src Lang.}'' and ``\textit{Tgt Lang.}'' denote the source and target languages for each dataset, respectively (En: English, Zh: Chinese and De: German).} 
  \label{table:syn_xls_datasets}
\end{table}

\vspace{0.5ex}
\noindent \textbf{Quality Controlling.} Since the translation results provided by MT services might contain flaws, En2ZhSum, Zh2EnSum and En2DeSum further use the round-trip translation (RTT) strategy to filter out low-quality samples. Specifically, given a monolingual document-summary pair $\langle D_{\textrm{src}},S_{\textrm{src}} \rangle$, the summary $S_{\textrm{src}}$ is first translated to a target langauge $S'_{\textrm{tgt}}$, and then $S'_{\textrm{tgt}}$ is translated back to the source language $S'_{\textrm{src}}$. Next, $\langle D_{src},S'_{tgt} \rangle$ will be retained as an XLS sample only if the \textrm{ROUGE} scores between $S_{src}$ and $S'_{src}$ exceed the pre-defined thresholds.
%
% In order to evaluate the model as correctly as possible
In addition, the translated summaries in the test set of En2ZhSum and Zh2EnSum are post-edited by human annotators to ensure the reliability of model evaluation.
% evaluate XLS models as correctly as possible.
% To correctly evaluate XLS models, the translated summaries in the test set of En2ZhSum and Zh2EnSum are post-edited by human annotator.

As for manually translated synthetic datasets, i.e., XSAMSum and XMediaSum, \citet{Wang2022ClidSumAB} design a quality control loop, where data reviewers and experts participate, to ensure the accuracy of the translation.

\vspace{0.5ex}
\noindent \textbf{Datasets Statistics.} Table~\ref{table:syn_xls_datasets} compares previous synthetic datasets in terms of the translation method, \modi{genre}, scale, source language and target language.
We conclude that:
% (1) the scale of datasets in the news domain is much larger than those in the dialogue domain.
(1) There is a trade-off between scale and quality. In line with MS, the scale of XLS datasets in news domain is much larger than others since news articles are convenient to collect. When faced with such large-scale datasets, it is expensive \modi{and} even impractical to manually translate or post-edit all their summaries. Thus, these datasets generally adopt automatic translation methods, causing limited quality.
(2) The XLS datasets in the dialogue domain are extremely challenging than those in the news domain. Besides the limited scale, the key information of one dialogue is often scattered and spanned multiple utterances, leading to low information density~\cite{Feng2021ASO}, which together with complex dialogue phenomena (e.g., coreference, repetition and interruption) \modi{makes} the task quite challenging~\cite{Wang2022ClidSumAB}.

\subsection{Multi-Lingual Website Datasets}
\label{subsec:mul_datasets}
In globalization process, online resources across different languages are overwhelmingly growing.
One reason is that many websites start to provide multi-lingual versions for their content to facilitate global users.
%
% \fimodi{These websites might contain a large number of monolingual document-summary pairs in multiple languages and also provide the alignment between parallel documents in different languages.}
\fimodi{Therefore, these websites might contain a large number of parallel documents in different languages.}
% to use their services, such as subscribing to digital news.
% Another way to establish XLS datasets
Some researchers try to utilize such resources to establish XLS datasets.

\vspace{0.5ex}
\noindent \textbf{Datasets Construction.} \citet{nguyen-daume-iii-2019-global} collect news articles from Global Voices website\footnote{\url{https://globalvoices.org/}} which reports and translates news about unheard voices across the globe. The translated news on this website is performed by volunteer translators. Each news article also links to its parallel articles in other languages, if available. Thus, it is convenient to obtain different language versions of \modi{an} article.
% Then, they collect two summaries for each language-specific article: one is the overall description on the HTML page and the other is written by crowdworkers. Lastly, the article in one language and its summary in a different language could constitute an XLS sample.
Then, they employ crowdworkers to write English summaries for hundreds of selected English articles. In this manner, the non-English articles together with the English summaries constitute the Global Voices XLS dataset\footnote{The Global Voices dataset contains gv-snippet and gv-crowd two subsets. The former cannot well meet the need of XLS due to its low quality~\cite{nguyen-daume-iii-2019-global}, thus we only introduce the gv-crowd subset.}.
Although this dataset utilizes online resources, the way to collect summaries (i.e., crowd-sourcing) limits its scale and directions (the target language must be English).

To alleviate the dilemma, WikiLingua~\cite{ladhak-etal-2020-wikilingua} collects multi-lingual guides from WikiHow\footnote{\url{https://www.wikihow.com/}}, where each step in a guide consists of a paragraph and the corresponding one-sentence summary.
Heuristically, the dataset combines paragraphs and one-sentence summaries of all the steps in one guide to create a monolingual article-summary pair.
With the help of hyperlinks between parallel guides in different languages, the article in one language and its summary in another one are easy to align. % and form an XLS sample
% Since no additional human annotation is required, the scale of WikiLingua is much larger than Global Voices.
In this way, WikiLingua collects articles and the corresponding summaries in 18 different languages, leading to 306 (18$\times$17) directions.
Similarly, \citet{perez-beltrachini-lapata-2021-models} construct XLS datasets from Wikipedia\footnote{\url{https://www.wikipedia.org/}}, a widely-used multi-lingual encyclopedia. In detail, the Wikipedia articles are typically organized into lead sections and bodies. They focus on 4 languages and pair lead sections with the corresponding bodies in different languages to construct XLS samples.
In the end, the collected samples form the XWikis dataset with 12 directions.

Moreover, \citet{Hasan2021CrossSumBE} construct CrossSum dataset by automatically aligning identical news articles written in different languages from the XL-Sum dataset~\cite{hasan-etal-2021-xl}. The multi-lingual news article-summary pairs in XL-Sum are collected from BBC website\footnote{\url{https://www.bbc.com/}}. As a result, CrossSum involves 45 languages and 1936 directions.

\vspace{0.5ex}
\noindent \textbf{Quality Controlling.} For manually annotated dataset, i.e., Global Voices, \citet{nguyen-daume-iii-2019-global} employ human evaluation to remove low-quality annotated summaries to ensure the quality.
For automatically collected datasets, i.e., WikiLingua and XWikis, they typically extract the desired content from the websites via heuristic matching rules to ensure the correctness.
As for automatically aligned dataset, i.e., CrossSum, \citet{Hasan2021CrossSumBE} adopt LaBSE~\cite{Feng2020LanguageagnosticBS} to encode all summaries from XL-Sum~\cite{hasan-etal-2021-xl}. Then, they align documents belonging to different languages based on the cosine similarity of corresponding summaries, and pre-define a minimum similarity score to reduce the number of incorrect alignments.

\vspace{0.5ex}
\noindent \textbf{Datasets Statistics.} Table~\ref{table:multi_web_xls_datasets} lists the key characteristics of the representative multi-lingual website datasets. It is worth noting that the number of XLS samples in each direction of the same dataset may be different since different articles might be available in different languages. Hence, we measure the overall scale of each dataset from its average, maximum and minimum number of XLS samples per direction, respectively.
We find that: (1) The scale of Global Voices is extremely less than other datasets due to the different methods to collect summaries. Specifically, WikiLingua, XWikis and XL-Sum (the basis of CrossSum) datasets automatically extract a huge amount of summaries from online resources via simple strategies rather than crowd-sourcing.
(2) CrossSum and WikiLingua involve more languages than others, and most of language pairs have intersectional articles, resulting in numerous cross-lingual directions.
% Among them, the largest and the least number of intersectional articles between two languages are 113k (English-Spanish) and 915 (Czech-Turkish), respectively.

\begin{table}[t]
  \centering
  \setlength{\belowcaptionskip}{-10pt}
  \resizebox{0.48\textwidth}{!}
  {
    \begin{tabular}{lcccc}
    \hline
     \multicolumn{1}{c}{\multirow{2}{*}{Dataset}} &  \multicolumn{1}{c}{\multirow{2}{*}{Domain}} & \multicolumn{1}{c}{\multirow{2}{*}{L}} & \multicolumn{1}{c}{\multirow{2}{*}{D}} & \multicolumn{1}{c}{\multirow{1}{*}{Scale}}  \\
     \multicolumn{1}{c}{} & \multicolumn{1}{c}{} & \multicolumn{1}{c}{} & \multicolumn{1}{c}{} & (avg / max / min)  \\ \hline
    % \multirow{5}{*}{\rotatebox[origin=c]{90}{synthetic} $\begin{dcases} \\ \\ \\ \\ \end{dcases}$}
    Global Voices       & News              & 15    & 14    & 208 / 487 / 75   \\
    CrossSum            & News              & 45    & 1936  & 845 / 45k / 1  \\
    WikiLingua          & Guides            & 18    & 306   & 18k / 113k / 915   \\
    XWikis              & Encyclopedia      & 4     & 12    & 214k / 469k / 52k  \\ \hline
    \end{tabular}
  }
  %  which are organized into synthetic datasets and multi-lingual website datasets
  \caption{Overview of representative multi-lingual website datasets. ``\textit{L}'' denotes the number of languages involved in each dataset. ``\textit{D}'' indicates the number of cross-lingual directions. ``\textit{Scale (avg/max/min)}'' calculates the average/maximum/minimum number of XLS samples per direction.} 
  \label{table:multi_web_xls_datasets}
\end{table}

\subsection{\modi{Discussion}}
\label{subsec:comparision_dataset}
\modi{According to the above review of large-scale XLS datasets, the approaches for building datasets are summarized as: (\uppercase\expandafter{\romannumeral1}) manually or (\uppercase\expandafter{\romannumeral2}) automatically translating the summaries of MS datasets; (\uppercase\expandafter{\romannumeral3}) automatically collecting documents as well as summaries from multi-lingual websites.}

\modi{Among them, approach \uppercase\expandafter{\romannumeral1} involves fewer noises than others since its translation and quality control are performed by professional translators rather than machine translation or volunteers. However, this approach is too labor-intensive and costly to build large-scale datasets. For instance, to control costs, XMediaSum~\cite{Wang2022ClidSumAB} only manually translates part of (\textasciitilde8.6\%) summaries of MediaSum~\cite{zhu-etal-2021-mediasum}.
Besides, Zh2EnSum and En2ZhSum~\cite{zhu-etal-2019-ncls} are automatically collected via approach \uppercase\expandafter{\romannumeral2}, and only their test sets have been manually corrected.
Therefore, despite the high quality of the constructed data, approach \uppercase\expandafter{\romannumeral1} is more suitable for building validation and test sets of large-scale XLS datasets rather than the whole datasets.
}

\modi{
Approaches \uppercase\expandafter{\romannumeral2} and \uppercase\expandafter{\romannumeral3} could be adopted to build whole XLS datasets.
%  Depending on whether the source and target languages are low-resource languages, respectively, we discuss them in the following situations:}
We discuss them in the following situations:}

\modi{(1) High-resource source languages $\Rightarrow$ high-resource target languages: This situation has been well studied in previous work, and most of the proposed XLS datasets focus on this situation. Both approaches \uppercase\expandafter{\romannumeral2} and \uppercase\expandafter{\romannumeral3} are useful to construct XLS datasets whose source and target languages are both high-resource languages.}

\modi{(2) High-resource source languages $\Rightarrow$ low-resource target languages: When the documents and summaries from XLS datasets are respectively in a high-resource language and a low-resource language, approach \uppercase\expandafter{\romannumeral3} loses its effectiveness. This is because, for a multi-lingual website, its content in a low-resource language is typically less than that in a high-resource language. As a result, the number of collected XLS samples involving low-resource languages is significantly limited. For example, WikiLingua~\cite{ladhak-etal-2020-wikilingua}, as a multi-lingual website dataset, contains 113.2k English$\Rightarrow$Spanish samples, but only 7.2k English$\Rightarrow$Czech samples. In this situation, approach \uppercase\expandafter{\romannumeral2} might be a possible way to collect a large number of samples. Note that the MT from a high-resource language to a low-resource language might involve more translation flaws than those between two high-resource languages. Thus, besides the RTT strategy, how to filter out the potential flaws is worthy to be further studied.}

\modi{(3) Low-resource source languages $\Rightarrow$ high- or low-resource target languages: If the source language is low-resource, there might be no MS dataset and enough website content in this language, leading to the failures of approaches \uppercase\expandafter{\romannumeral2} and \uppercase\expandafter{\romannumeral3}. Therefore, how to build datasets in this situation is still an open-ended problem, which needs to be explored in the future. As pointed by~\citet{feng-etal-2022-msamsum}, one straightforward approach is to automatically translate both documents and summaries from high-resource MS datasets. However, translating documents with hundreds of words might introduce substantial noise, especially when low-resource languages are involved. Thus, its practicality and reliability need more careful justification.}

%To provide a deeper understanding of current XLS datasets, we further compare the above two categories of datasets from the following aspects:

% \noindent \textbf{Quality.} 
% Building a good XLS system generally needs high-quality labeled samples. And it is not difficult to find that the quality of multi-lingual website datasets is higher than that of general-purpose (news domain) synthetic ones. This is because the translation in the former is typically performed by volunteer translators while the counterpart in the latter is processed by MT services or models. However, most current models only focus on the latter. Thus, we argue that the multi-lingual website datasets need more research attention.

%\noindent \textbf{Direction.} The cross-lingual directions in multi-lingual website datasets are more diverse than synthetic ones. In detail, all existing synthetic datasets are English-centered, i.e., English serves as either the source language or the target language. On the contrary, some of the multi-lingual website datasets (especially WikiLingua) provide comprehensive directions that are not limited to English.

\section{Pipeline Methods}
\label{sec:pipeline}
Early XLS work generally focuses on the pipeline methods whose main idea is decomposing XLS into MS and MT sub-tasks, and then accomplishing them step by step. These methods can be further divided into summarize-then-translate (Sum-Trans) and translate-then-summarize (Trans-Sum) types according to the finished order of sub-tasks.
For each type, we will systematically present previous methods. Besides, we compare these two types to provide deeper analyses.

\subsection{Sum-Trans}
\citet{Orasan2008EvaluationOA} utilize the Maximum Marginal Relevance (MMR) algorithm to summarize Romanian news, and then translate the summaries from Romanian to English via eTranslator MT service\footnote{\url{https://www.etranslator.ro/}}. Furthermore, \citet{wan-etal-2010-cross} find the translated summaries might fall into low readability due to the limited MT performance at that time. To alleviate this issue, they first use a trained SVM model~\cite{Cortes2004SupportVectorN} to predict the translation quality of each English sentence, where the model only leverages features in the English sentences. Then, they select sentences with high quality and informativeness to form summaries which are finally translated to Chinese by Google MT service\footnote{\url{https://cloud.google.com/translate}\label{fn:google}}.

\begin{figure*}[t]
\centerline{\includegraphics[width=1.00\textwidth]{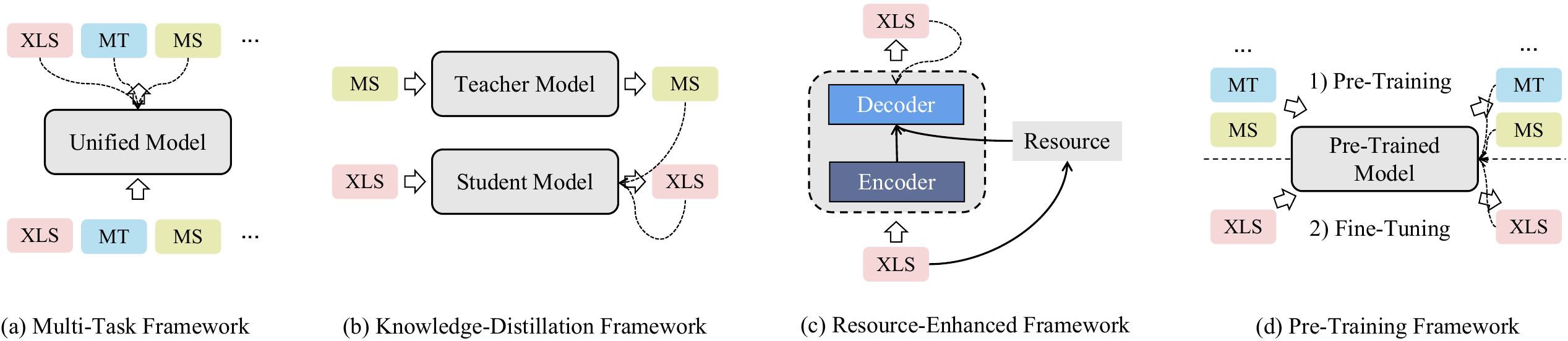}}
\caption{Overview of four end-to-end frameworks (best viewed in color). XLS: cross-lingual summarization; MT: machine translation; MS: monolingual summarization.
Dashed arrows indicate the supervised signals. Rimless colored blocks denote the input or output sequences of the corresponding tasks.
Note that the knowledge-distillation framework might contain more than one teacher model, and the auxiliary/pre-training tasks used in the multi-task/pre-training framework are not limited to MT and MS, here we omit these for simplicity. }
\label{fig:e2e}
% \vspace{-0.4cm}
\end{figure*} 

\subsection{Trans-Sum}

Compared with Sum-Trans, Trans-Sum attracts more research attention, and this type of pipeline method can be further classified into three sub-types depending on whether its summarizer is extractive, compressive or abstractive:
% (1) the extractive method selects complete sentences from the translated documents as summaries;
% (2) the compressive method first extracts key sentences from the translated documents, and further removes non-relevant or redundant words in the key sentences to obtain the final summaries;
% (3) the abstractive method generates new sentences as summaries, which are not limited to original words or phrases. Note that we do not classify the Sum-Trans approaches in the same manner since their summarizers are all extractive.

\begin{itemize}[leftmargin=*,topsep=0pt]
\setlength{\itemsep}{0pt}
\setlength{\parsep}{0pt}
\setlength{\parskip}{0pt}
\item The extractive method selects complete sentences from the translated documents as summaries.
\item The compressive method first extracts key sentences from the translated documents, and further removes non-relevant or redundant words in the key sentences to obtain the final summaries.
\item The abstractive method generates new sentences as summaries, which are not limited to original words or phrases.
\end{itemize}
Note that we do not classify the Sum-Trans approaches in the same manner since their summarizers are all extractive.

\vspace{0.5ex}
\noindent \textbf{Extractive Trans-Sum.}
\citet{Leuski2003CrosslingualCE} build a cross-lingual information delivery system which first translates Hindi documents to English via a statistical MT model and then selects important English sentences to form summaries. In this system, the summarizer only uses the document information from the target language side, which heavily depends on the MT results and might lead to flawed summaries. However, semantic information from both sides should be taken into account.

To this end, after translating English documents to Chinese, \citet{wan-2011-using} designs two graph-based summarizers (i.e., SimFusion and CoRank) which utilize bilingual information to output the final Chinese summaries: (\romannumeral1) the SimFusion summarizer first measures the saliency scores of Chinese sentences through combing the English-side and Chinese-side similarity, and then, the salient Chinese sentences constitute the final summaries;
% the SimFusion summarizer combines the English-side and Chinese-side similarity during measuring the saliency score of Chinese sentences; 
(\romannumeral2) the CoRank summarizer simultaneously ranks both English and Chinese sentences by incorporating mutual influences between them, and then, the top-ranking Chinese sentences are used to constitute summaries.

Later, \citet{Boudin2011AGA} translate documents from English to French, and then use SVM regression method to predict translation quality of each sentence based on bilingual features. Next, the crucial translated sentences are selected based on a modified PageRank algorithm~\cite{Page1999ThePC} considering the translation quality. Lastly, the redundant sentences are removed from the selected sentences to form the final summaries.
% based on the word diversity and sentence relevance.

\vspace{0.5ex}
\noindent \textbf{Compressive Trans-Sum.}
Inspired by phrase-based MT, \citet{yao-etal-2015-phrase} propose a compressive summarization method that simultaneously selects and compresses sentences. Specifically, the sentence selection is based on bilingual features, and the sentence compression is performed by removing the redundant or poorly translated phrases in one single sentence.
To further excavate the complementary information of similar sentences, \citet{Zhang2016AbstractiveCS} first parse bilingual documents into predicate-argument structures (PAS), and then produce summaries by fusing bilingual PAS structures. In this way, several salient PAS elements (concepts or facts) from different sentences can be merged into one summary sentence.
Similarly, \citet{Pontes2018CrossLanguageTS} take bilingual lexical chunks into account during measuring the sentence similarity and further compress sentences at both single- and multi-sentence levels.

% advent

\vspace{0.5ex}
\noindent \textbf{Abstractive Trans-Sum.}
With the emergence of large-scale synthetic XLS datasets~\cite{zhu-etal-2019-ncls}, researchers attempt to adopt the sequence-to-sequence models as summarizers in Trans-Sum methods.
Considering the translated documents might contain flaws, \citet{ouyang-etal-2019-robust} train \modi{an} abstractive summarizer (i.e., PGNet, \citealt{see-etal-2017-get}) on English pairs of a noisy document and a clean summary. In this manner, the summarizer could achieve good robustness, when summarizing the English documents which are translated from a low-resource language.
% when faced with translation flaws from a low-resource language to English, their summarizer achieves good robustness and performance.

\subsection{Sum-Trans vs. Trans-Sum}
We compare Sum-Trans and Trans-Sum in the following situations:
%(1) When using extractive or compressive summarizers, the summarizers of the Trans-Sum methods can benefit from bilingual documents while the counterpart of the Sum-Trans methods can only utilize the source-language documents. On the other hand, the Trans-Sum methods are less efficient since they need to translate the whole documents rather than a few summaries.
% (2) Besides the above discussion, when adopting abstractive summarizers, a large-scale summarization dataset is required to train the abstractive summarizers. Thus, the Trans-Sum methods are helpful if the source language is low-resource. In contrast, if the target language is low-resource in MS, the Sum-Trans methods are more useful.

\begin{itemize}[leftmargin=*,topsep=0pt]
\setlength{\itemsep}{0pt}
\setlength{\parsep}{0pt}
\setlength{\parskip}{0pt}
\item When using extractive or compressive summarizers, the summarizers of the Trans-Sum methods can benefit from bilingual documents while the counterpart of the Sum-Trans methods can only utilize the source-language documents.
\modi{Thus, the Trans-Sum methods typically achieve better performance than the Sum-Trans counterparts. For instance, on the manually translated DUC 2001 dataset, PBCS~\cite{yao-etal-2015-phrase}, as a Trans-Sum method, outperforms its Sum-Trans baseline by 8\%/8.4\%/10.4\% in terms of ROUGE-1/2/L.}
On the other hand, the Trans-Sum methods are less efficient since they need to translate the whole documents rather than a few summaries.
\item \modi{Apart from} the above discussion, when adopting abstractive summarizers, a large-scale \modi{MS} dataset is required to train the summarizers. \modi{It is also worth noting that the MS datasets in low-resource languages are much smaller than the MT counterparts~\cite{tiedemann-thottingal-2020-opus,hasan-etal-2021-xl}.} Thus, the Trans-Sum methods are helpful if the source language is low-resource. In contrast, if the target language is low-resource in MS, the Sum-Trans methods are more useful~\modi{\cite{ouyang-etal-2019-robust,ladhak-etal-2020-wikilingua}.}
\end{itemize}

% \romannumera
\section{End-to-End Methods}
\label{sec:e2e}
Though the pipeline method is intuitive, it 1) suffers from error propagation; 2) needs either a large corpus to train MT models or the monetary cost of paid MT services; 3) has a latency during inference.
% ~\cite{Ayana2018ZeroShotCN,duan-etal-2019-zero,zhu-etal-2019-ncls,zhu-etal-2020-attend,xu-etal-2020-mixed,Takase2020MultiTaskLF,dou-etal-2020-deep,cao-etal-2020-jointly,bai-etal-2021-cross,Ma2021DeltaLMEP,Nguyen2021ImprovingNC,chi-etal-2021-mt6,Liang2022AVH,Jiang2022ClueGraphSumLK,Wang2022ClidSumAB}
Thanks to the rapid development of neural networks, many end-to-end XLS models are proposed to alleviate the above issues.

In this section, we take stock of previous end-to-end XLS models and further divide them into four frameworks (cf., Figure~\ref{fig:e2e}): multi-task framework (\S~\ref{subsec:multi-task-framework}), knowledge-distillation framework (\S~\ref{subsec:kd-framework}), resource-enhanced framework (\S~\ref{subsec:re-framework}) and pre-training framework (\S~\ref{subsec:pt-framework}). For each framework, we will entirely introduce its core idea and corresponding models. At last, we discuss the pros and cons w.r.t each framework (\S~\ref{subsec:discussion}).

\subsection{Multi-Task Framework}
\label{subsec:multi-task-framework}
It is challenging for an end-to-end model to directly conduct XLS since it requires both the abilities to translate and summarize~\cite{cao-etal-2020-jointly}. As shown in Figure~\ref{fig:e2e}(a), many researchers use the related tasks (e.g., MT and MS) together with XLS to train unified models. In this way, XLS models could also benefit from the related tasks.

\citet{zhu-etal-2019-ncls} utilize a shared transformer encoder to encode the input sequences of both XLS and MT/MS. Then, two independent transformer decoders are used to conduct XLS and MT/MS, respectively. It is the first time to show the end-to-end method outperforms the pipeline ones.
Later, \citet{cao-etal-2020-jointly} use two encoder-decoder models to perform MS in the source and target languages, respectively.
Meanwhile, the source encoder and the target decoder jointly conduct XLS.
Then, two linear mappers are used to convert the context representation (i.e., the output of encoders) from the source to the target language and vice versa. In addition, two discriminators are adopted to discriminate between the encoded and mapped representations.
% Thereby, the overall model could jointly learn to both summarize documents as well as align representations between both languages.
Thereby, the overall model could jointly learn to summarize documents and align representations between both languages.

Although the above efforts design unified models in the multi-task framework, their decoders are independent for different tasks, leading to limitations in capturing the relationships among the multiple tasks.
To solve this problem, \citet{Takase2020MultiTaskLF} train a single encoder-decoder model on both MS, MT and XLS datasets. They prepend a special token at the input sequences to indicate which task is performed.
In addition, \citet{bai-etal-2021-cross} make the MS a prerequisite for XLS and propose MCLAS, a XLS model of single encoder-decoder architecture. For the given documents, MCLAS generates the sequential concatenation of the corresponding monolingual and cross-lingual summaries. In this way, the translation alignment is also implicit in the generation process, making MCLAS achieve great performance in XLS.
% Further, 
More recently, \citet{Liang2022AVH} utilize conditional variational auto-encoder (CVAE)~\cite{Sohn2015LearningSO} to capture the hierarchical relationship among MT, MS and XLS. Specifically, three variables are adopted in the proposed model to reconstruct the results of MT, MS and XLS, respectively. Besides, the used encoder and decoder are shared among all tasks, while the prior and recognition networks are independent to indicate the different tasks. Considering the limited XLS data in low-resource languages, \citet{bai-etal-2021-cross} and \citet{Liang2022AVH} also investigate XLS in the few-shot setting.
% After combining mBART~\cite{Liu2020MultilingualDP}, the model achieves the state-of-the-art performance on Zh2EnSum and En2ZhSum.

% \clearpage

\subsection{Knowledge-Distillation Framework}
\label{subsec:kd-framework}
% Before the emergence of large-scale XLS datasets, 
The original thought of knowledge distillation is distilling the knowledge in an ensemble of models (i.e., teacher models) into a single model (i.e., student model)~\cite{Hinton2015DistillingTK}.
Due to the close relationship between MT/MS and XLS, some researchers attempt to use MS or MT, or both models to teach the XLS model in the knowledge-distillation framework.
In this way, besides the XLS labels, the student model can also learn from the output or hidden state of the teacher models.

\citet{Ayana2018ZeroShotCN} utilize large-scale MS and MT corpora to train MS and MT models, respectively.
% Then, they investigate situations to teach the XLS student model: (1) using trained MS model as teacher model; (2) using trained MT model as teacher model and (3) using both trained MS and MT as teacher models. with bi-directional GRU architecture~\cite{Cho2014LearningPR}
%
Then, they use the trained MS or MT, or both models as the teacher models to teach the XLS student model.
Both the teacher and student models are bi-directional GRU models~\cite{Cho2014LearningPR}.
To let the student model mimic the output of the teacher model, the KL-divergence between the generation probabilities of these two models is used as the training objective.
% In addition, their experimental study demonstrates the superiority of considering both the MS and MT models as teacher models.
% change the architecture of both teacher and student models to transformer~\cite{vaswani2017attention}
% \footnote{An attention relay mechanism is used to dispose the attention weights.}
Later, \citet{duan-etal-2019-zero} implement transformer~\cite{vaswani2017attention} as the backbone of the MS teacher model and the XLS student model, and further train the student model with two objectives: (1) the cross-entropy between the generation distributions of these two models; (2) the Euclidean distance between the attention weights of both models.
% \footnote{The cross-attention weights of the student model are obtained by a designed relay mechanism.}
It is worth noting that both \citet{Ayana2018ZeroShotCN} and \citet{duan-etal-2019-zero} focus on zero-shot XLS due to the scarcity of XLS dataset at that time, while their training objectives do not include the XLS.

After the emergence of large-scale XLS datasets, \citet{Nguyen2021ImprovingNC} confirm the knowledge-distillation framework can also be adopted in rich-resource scenarios.
% Concretely
Specifically, \modi{they employ the transformer student and teacher models, and further propose a variant Sinkhorn divergence}, which together with the XLS objective \modi{supervises} the student XLS model.

\begin{CJK*}{UTF8}{gbsn}
\begin{table*}[t]
\centering
\resizebox{0.9\textwidth}{!}{%
\begin{tabular}{llc}
\toprule[1.5pt]
\multicolumn{1}{c}{Pre-Traing Task}                           & \multicolumn{1}{c}{Inputs}                                                     & \multicolumn{1}{c}{Targets}                                                    \\ \midrule[1.5pt]
Machine Translation (\texttt{MT})                & Everything that kills make me feel alive                   & 沉舟侧畔千帆过，病树前头万木春                                            \\ \hline
Cross-Lingual Summarization (\texttt{XLS})                        & Everything that kills make me feel alive                   & 向死而生                                                       \\ \hline
% Cross-Lingual Auto-Encoding (XAE)                     & that Everything make me kills alive feel                    & 沉舟侧畔千帆过，病树前头万木春                                            \\ \hline
Translation Span Corruption (\texttt{TSC})                 & \tabincell{l}{Everything [M1] make me [M2] alive. \\ 沉舟侧畔千帆过，病树前头万木春。} & [M1] that kills [M2] feel \\ \hline
Translation Pair Span Corruption (\texttt{TPSC})           & \tabincell{l}{Everything that [M1] me feel alive. \\ 沉舟侧畔[M2]过，病[M3]前头万木春。} & [M1] kills make [M2] 千帆\ [M3] 树 \\ \bottomrule[1.5pt]
\end{tabular}
}
% ``沉舟侧畔千帆过，病树前头万木春'' ``向死而生''
\caption{Examples of inputs and targets used by different cross-lingual pre-training tasks for the sentence ``Everything that kills make me feel alive'' with its Chinese translation and summarization. The randomly selected spans are replaced with unique mask tokens (i.e, [M1], [M2] and [M3]) in \texttt{TSC} and \texttt{TPSC}.}
\label{tab:examples}
\vspace{-0.3cm}
\end{table*}
\end{CJK*}

\subsection{Resource-Enhanced Framework}
\label{subsec:re-framework}

As shown in Figure~\ref{fig:e2e}(c),
the resource-enhanced framework utilizes additional resources to enrich the information of the input documents, and the generation probability of the output summaries is conditioned on both the encoded and enriched information.

\citet{zhu-etal-2020-attend} explore the translation pattern in XLS. In detail, they first encode the input documents in source language via a transformer encoder, and then obtain the translation distribution for the words of the input documents by the \texttt{fast-align} toolkit~\cite{dyer-etal-2013-simple}. Lastly, a transformer decoder is used to generate summaries in target language based on both its output distribution and the translation distributions. In this way, the extra bilingual alignment information helps the XLS model better learn the transformation from the source to the target language.
\citet{Jiang2022ClueGraphSumLK} utilize \texttt{TextRank} toolkit~\cite{mihalcea-tarau-2004-textrank} to extract key clues from input sequences, and then construct article graphs based on these clues via a designed algorithm. Next, they encode the clues and the article graphs by a clue encoder (with transformer encoder architecture) and a graph encoder (based on graph nerual networks), respectively.
Finally, a transformer decoder with two types of cross-attention (perform on the outputs of both clue and graph encoders) is adopted to generate final summaries. In addition, they consider the translation distribution used in \citet{zhu-etal-2020-attend} to further strength the proposed model.

\begin{table*}[t]
  \centering
  \resizebox{1.00\textwidth}{!}
  {
    \begin{tabular}{lllll}
\toprule[1.5pt]
\multicolumn{1}{c}{Model}                & Architecture & Training Objective           & Evaluation Direction                     & Evaluation Dataset                    \\ \midrule[1.5pt]
\multicolumn{5}{c}{Multi-Task Framework} \\ \midrule[1pt]
% TNCLS~\cite{zhu-etal-2019-ncls}         & Transformer  & XLS                    & En$\leftrightarrow$Zh                             & En2ZhSum, Zh2EnSum                    \\
CLS+MS~\cite{zhu-etal-2019-ncls}        & Transformer  & XLS+MS                 & En$\leftrightarrow$Zh                             & En2ZhSum, Zh2EnSum                    \\
CLS+MT~\cite{zhu-etal-2019-ncls}         & Transformer  & XLS+MT                 & En$\leftrightarrow$Zh                             & En2ZhSum, Zh2EnSum                    \\
\citet{cao-etal-2020-jointly}           & Transformer  & XLS+MS+\texttt{REC}                 & En$\leftrightarrow$Zh                             & Gigaword$^{\dagger}$, DUC2004$^{\dagger}$, En2ZhSum, Zh2EnSum \\
Transum~\cite{Takase2020MultiTaskLF}     & Transformer  & XLS+MS+MT              & Ar/Zh$\rightarrow$En, En$\rightarrow$Ja                      & DUC2004$^{\dagger}$, JAMUL$^{\dagger}$                        \\
MCLAS~\cite{bai-etal-2021-cross}        & Transformer  & XLS+MS                 & En$\leftrightarrow$Zh, En$\rightarrow$De                      & En2ZhSum, Zh2EnSum, En2DeSum          \\
VHM~\cite{Liang2022AVH}            & Transformer$^{*}$  & XLS+MS+MT              & En$\leftrightarrow$Zh                             & En2ZhSum, Zh2EnSum                    \\ \midrule[1pt]
\multicolumn{5}{c}{Knowledge-Distillation Framework} \\ \midrule[1pt]
MS teacher~\cite{Ayana2018ZeroShotCN}    & GRU          & XLS+\textit{KD} (MS)            & En$\rightarrow$Zh                                    & DUC2003$^{\dagger}$, DUC2004$^{\dagger}$                      \\
MT teacher~\cite{Ayana2018ZeroShotCN}    & GRU          & XLS+\textit{KD} (MT)            & En$\rightarrow$Zh                                    & DUC2003$^{\dagger}$, DUC2004$^{\dagger}$                      \\
MS+MT teachers~\cite{Ayana2018ZeroShotCN}     & GRU          & XLS+\textit{KD} (MS+MT)         & En$\rightarrow$Zh                                    & DUC2003$^{\dagger}$, DUC2004$^{\dagger}$                      \\
\citet{duan-etal-2019-zero}               & Transformer  & XLS+\textit{KD} (MS)            & Zh$\rightarrow$En                                    & Gigaword$^{\dagger}$, DUC2004$^{\dagger}$                               \\
\citet{Nguyen2021ImprovingNC}           & Transformer  & XLS+\textit{KD} (MS)            & En$\leftrightarrow$Zh, En$\leftrightarrow$Ja, En$\rightarrow$Ar/Vi & En2ZhSum, Zh2EnSum, WikiLingua        \\ \midrule[1pt]
\multicolumn{5}{c}{Resource-Enhanced Framework} \\ \midrule[1pt]
ATS~\cite{zhu-etal-2020-attend}          & Transformer$^{*}$ & XLS                    & En$\leftrightarrow$Zh                             & En2ZhSum, Zh2EnSum                    \\
GlueGraphSum~\cite{Jiang2022ClueGraphSumLK}      & Transformer$^{*}$ & XLS                    & En$\leftrightarrow$Zh                             & En2ZhSum, Zh2EnSum, CyEn2ZhSum$^{\ddagger}$        \\ \midrule[1pt]
\multicolumn{5}{c}{Pre-Training Framework} \\ \midrule[1pt]
\citet{xu-etal-2020-mixed}          & Transformer  & MLM+DAE+MS+MT+TSC  & En$\leftrightarrow$Zh                             & En2ZhSum, Zh2EnSum                    \\
\citet{dou-etal-2020-deep}       & Transformer  & XLS+MT+MS              & En$\rightarrow$Zh, En$\rightarrow$De                             & En2ZhSum, English-German$^{\ddagger}$              \\
mT6~\cite{chi-etal-2021-mt6}      & Transformer  & SC+TSC+PNAT            & Es/Ru/Tr/Vi$\rightarrow$En    & WikiLingua                            \\
$\Delta$LM~\cite{Ma2021DeltaLMEP}     & Transformer  & SC+TSC                 & Es/Ru/Tr/Vi$\rightarrow$En  & WikiLingua                            \\
m\textsc{Dial}BART~\cite{Wang2022ClidSumAB}        & Transformer  & AcI+UP+MS+MT           & En$\rightarrow$Zh, En$\rightarrow$De                             & XMediaSum40k                          \\ \bottomrule[1.5pt]
\end{tabular}
  }
  %  which are organized into synthetic datasets and multi-lingual website datasets
  % ``\textit{KD}'' denotes the knowledge distillation objectives, driven by the output of the corresponding teacher models, such as MS and MT.
  % ``\textit{KD}'' denotes the knowledge distillation objectives, followed by the training objectives of the corresponding teacher models.
  % ``\textit{CVAE-Trans}'' indicates the transformer encoder and decoder are adopted to CVAE.
  \caption{The summary of end-to-end XLS models. ``\textit{Transformer}'' means the vanilla transformer encoder-decoder architecture. $*$ denotes the variant architecture. ``\texttt{REC}'' represents the reconstruction objective, which is used to supervise the linear mappers in the model proposed by \citet{cao-etal-2020-jointly}. ``\textit{KD}'' denotes the knowledge distillation objectives, derived from the output or hidden state of the corresponding teacher models, such as MS and MT models. The ``\textit{Training Objective}'' of pre-trained models lists the pre-training objectives. Language nomenclature used in ``\textit{Evaluation Direction}'' is \href{https://en.wikipedia.org/wiki/List_of_ISO_639-1_codes}{ISO 639-1 codes}. $\dagger$ indicates the number of samples in the dataset is less than 2000. $\ddagger$ denotes unreleased datasets.} 
  \label{table:overview_e2e}
\end{table*}

\subsection{Pre-Training Framework}
\label{subsec:pt-framework}

The emergence of pre-trained models has brought NLP to a new era~\cite{Qiu2020PretrainedMF}.
% The pre-trained models typically first study the general language knowledge from the large-scale corpus and then focus on the specific task to learn the corresponding ability.
The pretrained models typically first learn the general representation from large-scale corpora, and then adapt to the specific task through fine-tuning.

More recently, the general multi-lingual pre-trained generative models have shown impressive performance on many multi-lingual NLP tasks.
% , such as mBART~\cite{Liu2020MultilingualDP} and mT5~\cite{Xue2021mT5AM},
\modi{For example, mBART~\cite{Liu2020MultilingualDP}, as a multi-lingual pre-trained model, is derived from BART~\cite{lewis-etal-2020-bart}. mBART is pre-trained with BART-style denoising objectives on a huge volume of unlabeled multi-lingual data.
mBART shows its superiority in MT originally~\cite{Liu2020MultilingualDP}, and \citet{Liang2022AVH} find it can also outperform many multi-task XLS models on large-scale XLS datasets through simply fine-tuning.
Later, mBART-50~\cite{Tang2020MultilingualTW} goes a step further and extends the language processing abilities of mBART from 25 languages to 50 languages.}
\modi{In addition to the BART-style pre-trained models, mT5~\cite{Xue2021mT5AM} is a multi-lingual T5~\cite{Raffel2020ExploringTL} model, which is pre-trained in 101 languages with the T5-style span corruption objective.}
\modi{Although great performance has been achieved}, these general pre-trained models only utilize the denoising or span corruption objectives in multiple languages without any cross-lingual supervision, resulting in the under-explored cross-lingual ability. 

To solve this problem, \citet{xu-etal-2020-mixed} propose a mix-lingual XLS model which is pre-trained with \modi{masked language model (MLM)}, \modi{denoising auto-encoder (DAE)}, MS, \modi{translation span corruption (\texttt{TSC})} and \texttt{MT} tasks\footnote{Typewriter font indicates the cross-lingual tasks.}. \fimodi{The \texttt{TSC} and \texttt{MT} pre-training samples are derived from OPUS English$\leftrightarrow$Chinese parallel corpus\footnote{\url{http://opus.nlpl.eu/}}}.
\citet{dou-etal-2020-deep} utilize \texttt{XLS}, \texttt{MT} and MS tasks to pre-train another XLS model. \fimodi{They leverage the English$\leftrightarrow$German/Chinese MT samples from WMT2014/WMT2017 dataset. For \texttt{XLS}, they pre-train the model on En2ZhSum and English-German datasets~\cite{dou-etal-2020-deep}.}
%
% \footnote{Typewriter font indicates the cross-lingual tasks and we will introduce these tasks in the next paragraph.}
\citet{Wang2022ClidSumAB} focus on dialogue-oriented XLS and extend mBART-50 with \modi{MS, \texttt{MT} and two dialogue-oriented pre-training objectives (i.e., action infilling and utterance permutation)} via the second pre-training stage \fimodi{on MediaSum and XMediaSum datasets}.
Note that \citet{xu-etal-2020-mixed}, \citet{dou-etal-2020-deep} and \citet{Wang2022ClidSumAB} only focus on XLS task.
\fimodi{The languages supported by these models are limited to a few specific ones.}

Furthermore, mT6~\cite{chi-etal-2021-mt6} and $\Delta$LM~\cite{Ma2021DeltaLMEP} are presented towards general cross-lingual abilities. In detail, \modi{\citet{chi-etal-2021-mt6} first presents three tasks, i.e., \texttt{MT}, \texttt{TSC} and translation pair span corruption (\texttt{TPSC}), to extend mT5, and then designs a PNAT decoding strategy to let the model separately decode each target span of \texttt{SC}-like pre-training tasks. Finally, \citet{chi-etal-2021-mt6} combine \texttt{SC}, \texttt{TSC} and PNAT to jointly train mT6 model.}
\fimodi{To support multiple languages, mT6 is pre-trained on CC-Net~\cite{wenzek-etal-2020-ccnet}, MultiUN~\cite{ziemski-etal-2016-united}, IIT Bombay~\cite{kunchukuttan-etal-2018-iit}, OPUS and WikiMatrix~\cite{schwenk-etal-2021-wikimatrix} corpora, covering a total of 94 languages.}
$\Delta$LM reuses the parameters of InfoXLM~\cite{chi-etal-2021-infoxlm} and further be trained with SC and \texttt{TSC} tasks \fimodi{on CC100~\cite{conneau-etal-2020-unsupervised}, CC-Net, Wikipedia dump, CCAligned~\cite{el-kishky-etal-2020-ccaligned} and OPUS corpora, including 100 languages}.
mT6 and $\Delta$LM have been demonstrated their superiority on WikiLingua (a large-scale XLS dataset).
Moreover, there are also some general cross-lingual pre-trained models that have not been evaluated in XLS, e.g., \textsc{Xnlg}~\cite{Chi2020CrossLingualNL} and \textsc{VeCo}~\cite{luo-etal-2021-veco}.

Table~\ref{tab:examples} shows the details of the above cross-lingual pre-training tasks.
% MT and XLS simply use the corresponding samples to pre-trained the model.
% Among them, XAE receives a noisy sentence and generates the translation of its original sentence.
%
\texttt{TSC} and \texttt{TPSC} predict the masked spans from a translation pair. The input sequence of \texttt{TSC} is only masked in one language while the counterpart of \texttt{TPSC} is masked in both languages.

\subsection{Discussion}
\label{subsec:discussion}
Table~\ref{table:overview_e2e} \modi{summarizes} all end-to-end XLS models.
% We summarize all end-to-end XLS models in Table~\ref{table:overview_e2e} and 
We conclude that all four frameworks resort to external resources to improve the XLS performance:
(1) The multi-task framework uses large-scale MS and MT corpora to help XLS. Though the multi-task learning is intuitive, its training strategy and weights of different task is non-trivial to determine.
(2) The knowledge-distillation framework is another way to utilize the large-scale MS and MT corpora. This framework is most suitable for zero-shot XLS since it could be supervised by the MS and MT teacher models without any XLS labels. Nevertheless, knowledge distillation often fails to live up to its name, transferring very limited knowledge from teacher to student~\cite{stanton2021does}. Thus, it should be verified more deeply in the rich-resource XLS.
(3) The resource-enhanced framework employs the off-the-shelf toolkits to enhance the representation of input documents. This framework significantly relaxes the dependence on external data, but it suffers from error propagation.
(4) The pre-training framework can benefit from both unlabeled and \modi{labeled} corpora. In detail, pre-trained models learn the general language knowledge from large-scale unlabeled data with self-supervised objectives. In order to improve the cross-lingual ability, they can resort to MT parallel corpus and design supervised signals. This framework absorbs more knowledge from more external corpora than others, leading to the \modi{promising} performance on XLS.

\modi{To give a deeper comparison of end-to-end XLS models, as shown in Table~\ref{table:experimental_results}, we organize a leaderboard with unified evaluation metrics, based on the released codes and generated results from representative published literature.
% we organize a leaderboard with unified evaluation metrics based on released codes as well as generated results from published work.
%
The models in the pre-training framework~\cite{Liu2020MultilingualDP,dou-etal-2020-deep,xu-etal-2020-mixed} generally outperform others. Besides, the pre-training framework could also serve other frameworks. For example, \citet{Liang2022AVH} utilize mBART weights as model initialization for VHM (i.e., mVHM), bringing decent gains compared with vanilla VHM.
Therefore, it is possible and valuable to combine the advantages of different frameworks, which is worthy of discussion in the future.
}

% (1) The multi-task framework uses large-scale MS and MT corpora to help XLS.
% Though the multi-task learning is intuitive, its training strategy and weights of different task is non-trivial to determine.
%  (2) The knowledge-distillation framework is another way to utilize the large-scale MS and MT corpora.
% This framework is most suitable for zero-shot XLS since it could be supervised by the MS and MT teacher models without any XLS labels.
% Nevertheless, knowledge distillation often fails to live up to its name, transferring very limited knowledge from teacher to student~\cite{stanton2021does}. Thus, it should be verified more deeply in the rich-resource XLS.
%  (3) The resource-enhanced framework employs the off-the-shelf toolkits to enhance the representation of input documents. This framework significantly relaxes the dependence on external data, but it suffers from error propagation.
% (4) The pre-training framework can benefit from both unlabeled and label corpora. In detail, pre-trained models learn the general language knowledge from large-scale unlabeled data with self-supervised objectives. To improve the cross-lingual ability, they can also resort to MT parallel corpus and carefully design supervised signals.
% This framework absorbs more knowledge from more external corpora than others, leading to the state-of-the-art performance on XLS.

% We only adopt an additional probabilistic bilingual lexicon instead of a large-scale parallel machine translation dataset, which significantly relaxes the model’s dependence on data. model’s dependence on data.

\begin{table}[t]
\centering
\setlength{\belowcaptionskip}{-10pt}
\resizebox{0.47\textwidth}{!}
{
\begin{tabular}{l||ccc||ccc}
\toprule[1.5pt]
\multicolumn{1}{c||}{\multirow{2}{*}{Model}} &  \multicolumn{3}{c||}{En2ZhSum} & \multicolumn{3}{c}{Zh2EnSum} \\
 & R-1      & R-2      & R-L     & R-1     & R-2    & R-L    \\ \midrule[1.5pt]
% TNCLS~\cite{zhu-etal-2019-ncls} & \heartsuit   & 36.82    & 18.72    & 33.20   & 38.85   & 21.93  & 35.05  \\
CLS+MS$^{\heartsuit \dagger}$~\cite{zhu-etal-2019-ncls}   & 38.25    & 20.20    & 34.76   & 40.34   & 22.65  & 36.39  \\
CLS+MT$^{\heartsuit \dagger}$~\cite{zhu-etal-2019-ncls}   & 40.23    & 22.32    & 36.59   & 40.25   & 22.58  & 36.21  \\
\citet{cao-etal-2020-jointly}$^{\heartsuit \dagger}$ & 38.12    & 16.76    & 33.86   & 40.97   & 23.20  & 36.96  \\
% MCLAS$^{\heartsuit \ddagger}$~\cite{bai-etal-2021-cross}   &          &          &         &         &        &        \\
VHM$^{\heartsuit *}$~\cite{Liang2022AVH}   & 40.98    & 23.07    & 37.12   & 41.36   & 24.64  & 37.15  \\
% \citet{Nguyen2021ImprovingNC}$^{\diamondsuit \ddagger}$   &          &          &         &         &        &        \\
ATS~\cite{zhu-etal-2020-attend}$^{\clubsuit \dagger}$         & 40.47    & 22.21    & 36.89   & 40.68   & 24.12  & 36.97  \\
mBART~\cite{Liu2020MultilingualDP}$^{\spadesuit \ddagger}$     & 41.55    & 23.27    & 37.22   & 43.61   & 25.14  & 38.79  \\
\citet{dou-etal-2020-deep}$^{\spadesuit *}$   & 42.83    & 23.30    & \textbf{39.29}   & -       & -      & -      \\
\citet{xu-etal-2020-mixed}$^{\spadesuit *}$     & \textbf{43.50}    & \textbf{25.41}    & 29.66   & 41.62   & 23.35  & 37.26  \\
mVHM~\cite{Liang2022AVH}$^{\heartsuit \spadesuit *}$  & 41.95    & 23.54    & 37.67   & \textbf{43.97}   & \textbf{25.61}  & \textbf{39.19}  \\ \bottomrule[1.5pt]
\end{tabular}
}
%  which are organized into synthetic datasets and multi-lingual website datasets
\caption{\modi{The leaderboard of end-to-end XLS models on En2ZhSum and Zh2EnSum datasets~\cite{zhu-etal-2019-ncls} in terms of ROUGE(R)-1/2/L~\cite{Lin2004ROUGEAP}. The evaluation scripts refer to~\citet{zhu-etal-2020-attend}.  $^{\heartsuit}$: multi-task framework; $^{\clubsuit}$: resource-enhanced framework; $^{\spadesuit}$: pre-training framework. $^{\dagger}$ indicates the results are obtained by evaluating output files provided by the authors; $^{\ddagger}$ denotes the results by running their released codes; $^{*}$ indicates the results are reported in the original papers which adopt the same evaluation scripts as~\citet{zhu-etal-2020-attend}.}}
\label{table:experimental_results}
\end{table}

\section{Prospects}
\label{sec:prospects}
% Presently, it is not completely clear whether the different construction methods of datasets have an influence on building XLS systems.

In this section, we discuss and suggest the following promising future directions, which meet actual application needs:

\vspace{0.5ex}
\noindent \modi{\textbf{The Essence of XLS.}} \modi{Unifying two abilities (i.e., translation and summarization abilities) in a single model is non-trivial~\cite{cao-etal-2020-jointly}. Even though the effectiveness of the state-of-the-art models has been proved, the essence of XLS remains unclear, especially (1) the hierarchical relationship between MT\&MS and XLS~\cite{Liang2022AVH}, and (2) the theoretical analysis for \textit{what makes MT\&MS help XLS?}}

\vspace{0.5ex}
\noindent \modi{\textbf{XLS Dataset with Low-Resource Languages.}} \modi{There are thousands of languages in the world and most of them are low-resource. Despite the practical significance, building high-quality and large-scale XLS datasets whose source or target language is low-resource remains challenging (c.f., Section~\ref{subsec:comparision_dataset}), and needs to be further explored in the future.}

\vspace{0.5ex}
\noindent \textbf{Unified XLS across \modi{Genres} and Domains.} As we described in Section~\ref{sec:datasets}, existing XLS datasets cover multiple \modi{genres} or domains, i.e., news, dialogue, guides and encyclopedia. The diversity across them is naturally promoting the need for unified XLS, instead of promoting the trend of devising unique models on individual \modi{genres} or domains. At present, the unified XLS is still under-explored, making us believe the urgent need for it.

\vspace{0.5ex}
\noindent \textbf{Controllable XLS.} \citet{Bai2021BridgingTG} integrate a compression rate to control how much information should be kept in the target language. If the compression rate is 100\%, XLS degrades to MT.
Thus, the continuous variable unifies XLS and MT tasks. In this manner, a new research view is introduced to leverage MT to help XLS.
In addition, controlling some other attributes of the target summary may be useful in real applications, such as entity-centric XLS and aspect-based XLS.

\vspace{0.5ex}
\noindent \textbf{Low-Resource XLS.} Most languages in the world are low-resource, which makes large-scale parallel datasets across these languages rare and expensive. Hence, low-resource XLS is more realistic.
Nevertheless, current work has not well investigate and explore this situation.
Recently, prompt-based learning has become a new paradigm in NLP~\cite{Liu2021PretrainPA}. With the help of the well-designed prompting function, a pre-trained model is able to perform few-shot or even zero-shot learning.
Future work can adopt prompt-based learning to deal with the low-resource XLS.

\vspace{0.5ex}
\noindent \textbf{Triangular XLS.} Following triangular MT, triangular XLS is a special case of low-resource XLS where the language pair of interest has limited parallel data, but both languages have abundant parallel data with a pivot language. This situation typically appears in multi-lingual website datasets (a category of XLS datasets, cf., \S~\ref{subsec:mul_datasets}), because their documents are usually centered in English and then translated into other languages to facilitate global users. Hence, English acts as the pivot language. How to exploit such abundant parallel data to improve the XLS of interest language pairs remains challenging.

\vspace{0.5ex}
\noindent \textbf{Many-to-Many XLS.} Most previous work trains XLS models separately in each cross-lingual direction. In this way, the knowledge of XLS cannot be transferred among different directions. Besides, a trained model can only perform in a single direction, resulting in limited usage.
To solve this problem, \citet{Hasan2021CrossSumBE} jointly fine-tune mT5 in multiple directions.
% Cao_Wan_Yao_Yu_2020
At last, the fine-tuned model can perform in arbitrary even unseen directions, which is named many-to-many XLS.
Future work can focus on designing robust and effective training strategies for many-to-many XLS.

\vspace{0.5ex}
\noindent \textbf{Long Document XLS.} Recently, long document MS has attracted wide research attention~\cite{cohan-etal-2018-discourse,sharma-etal-2019-bigpatent,Wang2021SportsSum20GH,Wang2022KnowledgeES}.
Long document XLS is also important in real scenes, e.g., facilitating researchers to access the arguments of scientific papers in foreign languages. Nevertheless, this direction has not been noticed by previous work.
Interestingly, we find many non-English scientific papers have the corresponding English abstracts due to the regulations of publishers.
For example, many Chinese academic journals require researchers to write abstracts in both Chinese and English.
This might be a feasible method to construct long document XLS datasets.
We hope future work can promote this direction.
% It is challenging to deal with long documents for pre-trained models due to the time and memory complexity of training.
% Long documents pose challenges for pre-trained models due to the time and memory complexity of training.
% To the best of our knowledge, there is no multi-lingual pre-trained models focus on long documents

\vspace{0.5ex}
\noindent \textbf{Multi-Document XLS.} Previous multi-document XLS work~\cite{Orasan2008EvaluationOA,Boudin2011AGA,Zhang2016AbstractiveCS} only utilizes statistical features to build pipeline systems, and further performs on early XLS datasets.
The multi-document XLS is also worthy of discussion in the pre-trained model era.

\vspace{0.5ex}
\noindent \textbf{Multi-Modal XLS.} With the increasing of multimedia data on the internet, some researchers put their effort into multi-modal summarization~\cite{zhu-etal-2018-msmo,sanabria18how2,ijcai2018-577,li-etal-2020-vmsmo,fu-etal-2021-mm}, where the input of summarization systems is a document together with images or videos. Nevertheless, existing multi-modal summarization work only focuses on the monolingual scene and ignores cross-lingual ones. We hope future work could light up multi-modal XLS.

\vspace{0.5ex}
\noindent \textbf{Evaluation Metrics.} Developing evaluation metrics for XLS is still an open problem that needs to be further studied. Current XLS metrics typically inherit from MS. However, different from MS, the XLS samples consist of $\langle$source document, (target document), source summary, target summary$\rangle$. Besides the target summary, how to apply other information to assess the summary quality would be an interesting point for further study.

\vspace{0.5ex}
\noindent \textbf{Others.} Considering current XLS research is still in the preliminary stage, many research points of MS are missing in XLS, such as the factual inconsistency and hallucination problems.
% Moreover, to the best of our knowledge, there is no large-scale multi-document XLS dataset.
%
These directions are also worthy to be deeply explored in further work.

\section{Conclusion}
\label{sec:conclusion}
In this paper, we present the first comprehensive survey of current research efforts on XLS.
We systematically summarize existing XLS datasets and methods, highlight their characteristics and compare them with each other to provide deeper analyses. In addition, we give multiple perspective directions to facilitate further research on XLS.
We hope that this XLS survey could provide a clear picture of this topic and boost the development of the current XLS technologies.

\section*{Acknowledgements}
We would like to thank anonymous reviewers for their suggestions and comments.
This research is supported by the National Key Research and Development Project (No. 2020AAA0109302), the National Natural Science Foundation of China (No. 62072323, 62102276), Shanghai Science and Technology Innovation Action Plan (No. 19-511120400), Shanghai Municipal Science and Technology Major Project (No. 2021SHZDZX01-03), the Natural Science Foundation of Jiangsu Province (Grant No. BK20210705), the Natural Science Foundation of Educational Commission of Jiangsu Province, China (Grant No. 21KJD52-0005) and the Priority Academic Program Development of Jiangsu Higher Education Institutions.

%  the Natural Science Foundation of Jiangsu Province (Grant No. BK20210705), the Natural Science Foundation of Educational Commission of Jiangsu Province, China (Grant No. 21KJD520005) and the Priority Academic Program Development of Jiangsu Higher Education Institutions. 

\bibliography{tacl2021}
\bibliographystyle{acl_natbib}

\end{document}